\pdfoutput=1
\PassOptionsToPackage{table,usenames,dvipsnames,svgnames}{xcolor}
\documentclass[11pt]{article}

\usepackage[final]{coling}

\usepackage{times}
\usepackage{latexsym}

\usepackage[T1]{fontenc}

\usepackage{makecell}
\usepackage{booktabs}
\usepackage{CJKutf8}
\usepackage{multirow}
\usepackage{graphicx}
\usepackage{pifont}
\usepackage{setspace}
\usepackage[most]{tcolorbox}

\usepackage[utf8]{inputenc}
\usepackage[linesnumbered,lined,ruled,commentsnumbered]{algorithm2e}
\usepackage{bm}

\usepackage{microtype}

\usepackage{inconsolata}

\usepackage{graphicx}

\definecolor{amaranth}{rgb}{0.9, 0.17, 0.31}
\definecolor{kellygreen}{rgb}{77, 186, 23}
\definecolor{azure}{rgb}{0.0, 0.5, 1.0}
\definecolor{gred}{rgb}{0.9, 0.17, 0.31}
\definecolor{gblue}{rgb}{0.0, 0.5, 1.0}
\definecolor{gyellow}{RGB}{244,180,0}
\definecolor{ggreen}{rgb}{0.3, 0.73, 0.09}
\definecolor{ggrey}{RGB}{115,115,115}

\newcommand{\cmark}{{\color[HTML]{00715E}{\ding{51}}}}
\newcommand{\xmark}{{\color[HTML]{961C26}{\ding{55}}}}
\newcommand{\error}[1]{\textcolor{gred}{\textbf{#1}}}
\newcommand{\fph}[1]{\textcolor{gblue}{\textbf{#1}}}

\newcommand{\start}[1]{\vspace{.6mm}\noindent{{\bf #1}\ }}

\newcommand{\cls}{\texttt{[CLS]}\xspace}
\newcommand{\sep}{\texttt{[SEP]}\xspace}

\title{Two-stage Incomplete Utterance Rewriting on Editing Operation}

\author{Zhiyu Cao, Peifeng Li\thanks{ \ \ Corresponding author}, Qiaoming Zhu, Yaxin Fan \\
        School of Computer Science and Technology, Soochow University, Suzhou, China  \\
        \texttt{zycao18@stu.suda.edu.cn, yxfansuda@stu.suda.edu.cn} \\
        \texttt{\{pfli, qmzhu\}@suda.edu.cn}
        }

\begin{document}
\maketitle
\begin{abstract}
Previous work on Incomplete Utterance Rewriting (IUR) has primarily focused on generating rewritten utterances based solely on dialogue context, ignoring the widespread phenomenon of coreference and ellipsis in dialogues. To address this issue, we propose a novel framework called TEO (\emph{Two-stage approach on Editing Operation}) for IUR, in which the first stage generates editing operations and the second stage rewrites incomplete utterances utilizing the generated editing operations and the dialogue context. Furthermore, an adversarial perturbation strategy is proposed to mitigate cascading errors and exposure bias caused by the inconsistency between training and inference in the second stage. Experimental results on three IUR datasets show that our TEO outperforms the SOTA models significantly.
\end{abstract}

\section{Introduction}

Dialogue understanding (e.g., dialogue generation, dialogue sentiment analysis, and intent recognition) often 
suffers from ellipsis and coreference, because people often omit certain information or use pronoun in utterances for the sake of convenience in real-world conversations. Incomplete Utterance Rewriting (IUR) is to  rewrite incomplete utterances more specific and direct, which is beneficial for  many downstream dialogue understanding tasks, such as conversational dense retrieval \cite{qian-dou-2022-explicit} and dialogue summarization \cite{fang-etal-2022-spoken}.
Actually, IUR can be specifically categorized into coreference and ellipsis resolution. As shown in Table~\ref{rewrite-examples}, the incomplete utterance $u_3$ uses the pronoun ``\emph{he}'' to represent ``\emph{Ben Affleck}'' and omits ``\emph{as Batman}''. The rewritten utterance $u_3^{\prime}$ ``\emph{It is Ben Affleck who acted as Batman}'' is more informative and direct than the original incomplete utterance ``\emph{It is he who acted}''.

Although previous methods have achieved great success, they only regarded IUR as a generation or sequence labeling task, which did not explicitly consider the two different operations of replacement (for coreference) and insertion (for ellipsis)  \citep{hao-etal-2021-rast, chen-2023-incomplete,li-etal-2023-incomplete, 10384316}. 
Coreference resolution typically only involves the replacement of a single span (most of them are entities, e.g., \emph{``Ben Affleck''}), while ellipsis resolution might insert multiple discontinuous spans (e.g., \emph{``as''} and \emph{``Batman''}) from dialogue history at the same position in an incomplete utterance. 
Therefore, previous work treated coreference and ellipsis as a whole, which failed to account for the fundamental distinction between coreference and ellipsis. This resulted in the generation of utterances containing erroneous tokens or structures.

\begin{table}[t]
\centering
\resizebox{\linewidth}{!}{
\begin{tabular}{cc}
\toprule
\textbf{Speaker(turn)} & \textbf{Utterance}\\
\midrule
$\mathbf{Speaker_1(u_1)}$ & \makecell[c]{I think Batman is very handsome.}  \\
\midrule
$\mathbf{Speaker_2(u_2)}$ & \makecell[c]{The poster looks a bit like Ben Affleck.} \\
\midrule
$\mathbf{Speaker_1(u_3)}$ & \makecell[c]{It is \error{he} who acted. (Incomplete utterance)}  \\
\midrule
$\mathbf{Speaker_1(u_3^{\prime})}$ & \makecell[c]{It is \error{Ben Affleck} who acted \fph{as Batman}.}  \\
\bottomrule
\end{tabular}
}
\caption{
An example of IUR, where \error{red} and \fph{blue} color indicate coreference and ellipsis, respectively. The first two utterances u$_1$ and u$_2$ are dialogue history, the third u$_3$ is an incomplete utterance to be rewritten, and the fourth 
u$_3^{\prime}$ is the rewritten utterance. 
}
\label{rewrite-examples}
\vspace{-0.6cm}
\end{table}

To  address the aforementioned issues, we draw inspiration from the process of human article editing. When editing an article, humans typically read the entire text first, identify each area that requires insertion, deletion or replacement, and then reread the entire article to assess the reasonableness of each change. This prompts the question of \emph{whether it is possible to explicitly introduce editing operations into incomplete utterance rewriting.}

We instantiate this idea by adopting a similar pipeline framework with a generation model to simulate human article editing. Specifically, we propose a novel framework called TEO (\emph{Two-stage approach on Editing Operation}). 
We first use the editing operations as the pivot in the rewriting process, and train a model to generate  editing operations in the first stage. The second stage is responsible for generating rewritten utterances based on the dialogue context and the editing operations. Furthermore, 
to mitigate the exposure bias caused by inconsistency between training and inference in the second stage, we propose three perturbation methods of editing operations to improve robustness. The experimental results on three IUR datasets show that our TEO significantly outperforms the SOTA models. In summary, our two-stage generation framework TEO has three advantages: 
\begin{itemize}
\vspace{-5pt}
 \item The editing operations are taken as the pivot, with both insertion and replacement operations explicitly considered in order to address the issues of coreference and ellipsis.
\vspace{-5pt}
\item The local editing operations are first generated, and then the global dialogue context information and the editing operations are used to generate the final rewritten utterances. This enables the TEO to capture more fine-grained information. 
\vspace{-5pt}
\item An adversarial perturbation strategy is proposed for editing operations that can mitigate the occurrence of cascading errors and exposure bias caused by inconsistency between training and inference.
\end{itemize}

\begin{figure*}[ht]
\begin{center}
 \includegraphics[width=1\linewidth]{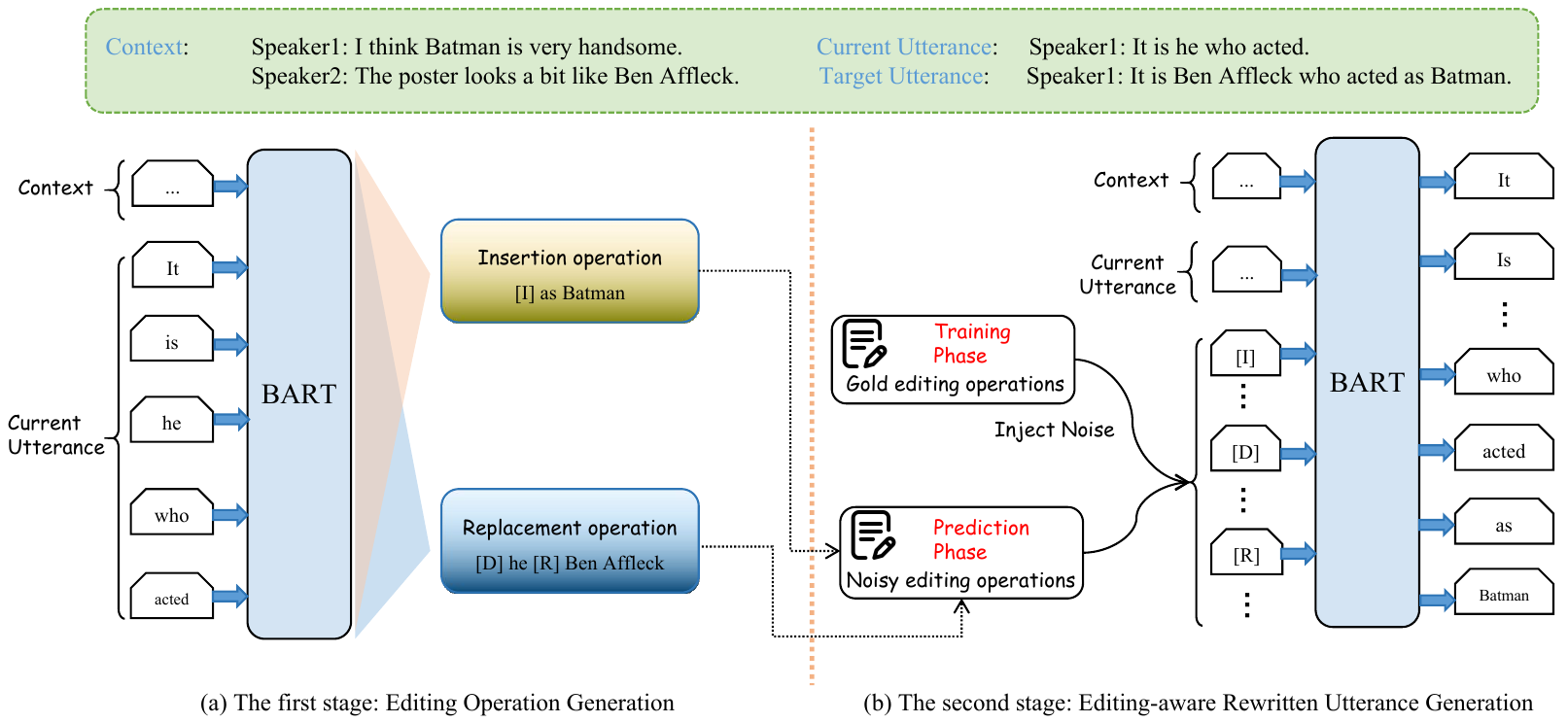}
 \caption{An overview of our framework TEO, which includes two stages Editing Operation Generation and Editing-aware Rewritten Utterance Generation.}
 \label{fig:method}
\end{center}
\vspace{-0.4cm}
\end{figure*}

\section{Related Work}

Research on IUR can be mainly divided into two types: generation methods \citep{DBLP:conf/aaai/HuangLZ021,inoue-etal-2022-enhance, li-etal-2023-incomplete} and sequence labeling methods \citep{liu-etal-2020-incomplete, DBLP:conf/aaai/JinSJ0G22, si-etal-2022-mining, chen-2023-incomplete, du-etal-2023-multi, li-etal-2023-well, 10384316}. 

Most previous studies did not explicitly consider coreference resolution and ellipsis resolution. The initial research on IUR predominantly employed generation methodologies. For example, 
\citet{DBLP:conf/aaai/HuangLZ021} first employed a tagger to predict the rewriting labels and then utilized an autoregressive with copy mechanism for generating the rewritten utterances.
\citet{inoue-etal-2022-enhance} jointly trained two tasks: selecting key words and generating rewritten utterances. 

Subsequent studies have indicated that the source and target utterances exhibit similar structural characteristics. Consequently, numerous subsequent studies have proposed sequence labelling methods. For example, 
\citet{liu-etal-2020-incomplete} regarded incomplete utterance rewriting as predicting word-level editing matrix. 
To address the issue of inserting multiple spans at one location, \citet{chen-2023-incomplete} directly selected spans from the context to form complete utterances. \citet{du-etal-2023-multi} incorporated sentence-level semantic relations between dialogue context and incomplete utterance. \citet{li-etal-2023-well} introduced the MLP architecture to mine the correlation between the contextual utterances and the rewritten utterance to obtain the editing matrix. \citet{10384316} paired spans and labeling the action types between spans.

Only a few works on IUR  \citep{si-etal-2022-mining, li-etal-2023-incomplete} explicitly considered coreference and ellipsis resolution. \citet{si-etal-2022-mining} inserted markers into incomplete utterances to represent coreference and ellipsis through manually designed rules. However, these rules cannot cover all situations. \citet{li-etal-2023-incomplete} first predicted the positions of insertion and replacement, and then filled these positions. However, they replaced the spans to be replaced with \texttt{[MASK]}, resulting in the loss of span information before replacement.

\section{Methodology}

\subsection{Task Definition}
Let each piece of data be defined as $\{Hist,U_n,Y\}$, where $Hist=\{U_1,U_2,...U_{n-1}\}$ is the dialogue history including $n-1$ utterances and $U_n$ is the incomplete utterance that requires rewriting. The rewritten utterance of $U_n$ is denoted as $Y$, where $Y$ keeps the semantics of $U_n$ unchanged and complements the coreferential and omitted information in it. The goal of IUR is to learn a mapping $P(T|Hist, U_n)$ satisfying $Y = \mathop{\arg\max}\limits_{T} P(T|Hist, U_n)$.

\subsection{Overview}
We adopt a text generation approach to the IUR task. In contrast to previous research, which has treated IUR as a single task, we decompose it into two subtasks: editing operation generation and editing-aware rewritten utterance generation  by using the editing operations $\mathcal{E}$ as the pivot. The objective of the overall training process is to maximize
\begin{equation}
\label{equation:total_tgt}
    \small P(Y|Hist, U_n) = P(\mathcal{E}|Hist, U_n) P(Y|Hist, U_n, \mathcal{E}),
\end{equation}
with the goal of maximizing $P(\mathcal{E}|Hist, U_n)$ in the first stage and maximizing $P(Y|Hist, U_n, \mathcal{E})$ in the second stage, where $\mathcal{E}$ is the predicted editing operations in the first stage.

Our framework TEO is shown in Figure \ref{fig:method}. In the first stage \emph{Editing Operation Generation}, we generate the editing operations based on the dialogue history and the incomplete utterance that needs to be rewritten. 
Subsequently, in the second stage \emph{Editing-aware Rewritten Utterance Generation}, we generate the rewritten utterance based on the editing operations and dialogue context. 

\subsection{Editing Operation Generation}
In the process of text editing, humans primarily engage in the insertion, deletion, and replacement of tokens. In the first stage, the objective is to generate editing operations to guide IUR. As previously stated, IUR primarily aims to address the coreference and ellipsis issues in utterances. The coreference resolution is typically achieved through replacement operations, while the ellipsis resolution is typically achieved through insertion operations. Consequently, two editing operations have been defined for IUR: insertion and replacement. 

To express the semantics of the editing operations in a more concise manner, it is not necessary to use natural language descriptions to represent insertion and replacement operations. Instead, three types of markers are employed for insertion and replacement. Specifically, we use ``$[I]$ $ tk$'' to represent an insertion operation where ``$[I]$'' refers to the insertion action and ``$tk$'' refers to the inserted tokens. The replacement operation is indicated by the following sequence: ``$[D]$ $tk_1$ $[R]$ $tk_2$''. Here, ``$[D]$'' and ``$[R]$'' refer to the deletion and replacement action, respectively, and the tokens ``$tk_1$'' are replaced by ``$tk_2$''.

By doing so, we can obtain the structured editing operations and denote it as $\mathcal{E}$, where each of the editing operations is ordered according to its position in the original utterance. As shown in Figure \ref{fig:method}, by comparing the incomplete utterance \emph{``It is he who acted''} with the rewritten utterance \emph{``It is Ben Affleck who acted as Batman''}, we can obtain the set of editing operations as ``\emph{[D] he [R] Ben Affleck [I] as Batman}''. 

To train a generation model to generate editing operations, we use the incomplete utterances and the rewritten utterances in the training set to create editing operations.  Specifically, we first compare the incomplete utterance $U_n$ and the rewritten utterance $Y$ to find their longest common subsequence, denoting as $\mathcal{L}$. For those tokens that appear in $U_n$ but not in $\mathcal{L}$, we mark them as \emph{deletion}. For those tokens that appear in  $Y$ but not in $\mathcal{L}$, we mark them as \emph{insertion}. Finally, we obtain a set of deletion and insertion operations. For each deletion and insertion operation with the same context, we label it as \emph{replacement}, resulting in a set of replacement operations. By using the above methods, we can obtain the insertion operation set $\mathcal{I}$ and the replacement operation set $\mathcal{R}$. For each incomplete utterance, our goal of editing operation generation is to generate the set of editing operations as follows,
\begin{equation}
    \small T(\mathcal{I},\mathcal{R}) = \text{[I]}I_1 ... \text{[I]}I_i...\text{[D]}D_1\text{[R]}R_1...\text{[D]}D_j\text{[R]}R_j ,
    \label{equation:edit-target}
\end{equation}
where $I_i$ represents the $i$-th span to be inserted, and $D_j$ and $R_j$ represent the $j$-th span to be replaced and the span after replacement, respectively.

We use BART \citep{lewis-etal-2020-bart} as our generation model and concatenate the dialogue history $Hist$ and the incomplete utterance $U_n$ as input. The output is the set of editing operations $\mathcal{E}$ of length $l$. The training objective of the editing operation generation is to minimize the negative log-likelihood loss function $ Loss_{edit}$ as follows,
\begin{equation}
\label{equation:edit-loss}
    \small Loss_{edit} = -\log \sum_{i=1}^{l} P(\mathcal{E}_i|\mathcal{E}_{<i}, Hist, U_n, \theta_1),
\end{equation}
where $\theta_1$ is the parameters of the model in the first stage.

\subsection{Editing-aware Rewritten Utterance Generation}\label{sec:editgen}

Our editing process in the first stage can be seen as a process of proposing insertions and replacements. Due to the errors in the predicted editing operations of the first stage and the unknown positions for the insertion operations, we combine global dialogue information with the preliminary generated editing operations to generate the final rewritten utterances in the second stage. We hope that the model can review and refine the generated editing operations based on contextual information, in order to obtain correct rewritten utterances. Specifically, we train the model to generate the final rewritten utterances based on the dialogue history, the utterances to be rewritten, and the editing operations. Moreover, to construct editing-aware context data,  we use the correct editing operations during the training stage and use the noisy editing operations generated by the first stage during the prediction stage.  

The exposure bias caused by the inconsistency in the training and inference stages can affect the effectiveness of the model, leading to overfitting of correct editing operations. Here we adopt three editing operation perturbation methods to alleviate exposure bias. Specifically, we add three types of perturbations to the editing operations during training to improve the robustness of the model, namely random replacement, random deletion and random insertion. 

\begin{algorithm}
\small
\SetKwData{e}{$e^*$}\SetKwData{ES}{$ES$}
\SetKwFunction{isadj}{IsAdj}
\SetKwFunction{addedge}{AddEdge}
\SetKwInOut{Input}{input}\SetKwInOut{Output}{output}
\Input{Editing operations $\mathcal{T}$; Perturbation probability $prob_p$; Span replacement probability $prob_r$}
\Output{Perturbed editing operations $E_{p}$}
$E_{p}\leftarrow \{\}$\;
\ForEach {$e$ in $\mathcal{T}$}
{
Draw $prob_1$, $prob_2$ from Uniform(0, 1)\;
\tcp{Add noise to editing operations with a probability of $prob_p$}
\eIf{$prob_1<=prob_p$} {
    \If{$prob_2<=prob_r$} {
        $e.text \leftarrow$ A random span sampled from the dialogue history\;
        $E_{p} \leftarrow E_{p} \cup \{e\}$
    }
}
{
    $E_{p} \leftarrow E_{p} \cup \{e\}$
}
}
Draw $prob_3$ from Uniform(0, 1)\;
\If{$prob_3<=prob_p$} {
    origin\_span $\leftarrow$ A random span sampled from the incomplete utterance\;
    new\_span $\leftarrow$ A random span sampled from the dialogue history\;
    candidates=[``[I] new\_span'', ``[D] origin\_span [R] new\_span'']\;
    $E_{p} \leftarrow E_{p} \cup$ \{the operation randomly sampled from the candidates\}
}
return $E_{p}$
\caption{Perturbation Strategy}
\label{perturbation}

\end{algorithm}

The probability of perturbations is denoted by $prob_p$. For gold editing operations, the probability of randomly replacing the span of the operation is $prob_r$, while the probability of randomly deleting it is $1-prob_r$.  For each editing operation $e$, we first sample two probability values $prob_1$ and $prob_2$ from a uniform distribution. If $prob_1 \leq prob_p$, we perform random deletion or random replacement as follows: if $prob_2 \leq prob_r$, we randomly replace the text (corresponding span) with a random span sampled from dialogue (Line 6-9); otherwise, we delete this editing operation (i.e., do not insert into $E_p$).
Taking the editing operation ``\emph{[D] he [R] Ben Affleck}'' as an example, we use ``\emph{Batman}'' to replace ``\emph{Ben Affleck}'' and form a new editing operation ``\emph{[D] he [R] Batman}''.
Then, we randomly sample a probability value $prob_3$ from a uniform distribution. if $prob_3 \leq prob_p$, we perform a random insertion (Line 15-20), including an insertion or a replacement operation. 
Taking the Figure \ref{fig:method} as an example, we insert an insertion operation ``\emph{[I] Batman}'' or a replacement operation ``\emph{[D] He [R] Batman}'' to $E_p$, resulting in an adversarial sample.

We also use BART as the generation model and concatenate the history ${Hist}$, the incomplete utterance ${U_n}$ and the editing operations $\mathcal{E}$ as input, i.e., $\{\cls{Hist}\sep {U_n}\sep{\mathcal{E}}\sep\}$, where $\cls$ and $\sep$ represent the special tokens in BART. The output is the rewritten utterance $Y=\{y_1,...y_i,...y_m \}$ where $y_i$ is the $i$-th token. The training objective of the editing-aware rewritten utterance generation is to minimize the negative log-likelihood loss function $Loss_{uttr}$ as follows,
\begin{equation}
\label{equation:uttr-loss}
    \small Loss_{uttr} = -\log \sum_{i=1}^{m} P(y_i|y_{<i}, Hist, U_n, \mathcal{E},\theta_2),
\end{equation}
where  $\theta_2$ is the parameters of the model in the second stage. We initialize the model using the weights of the model trained in the first stage. 

\section{Experimentation}
\subsection{Experimental Settings}

\begin{table}[]
\centering
\small
\setlength{\tabcolsep}{2.8mm}
\begin{tabular}{lccccccccccccc}
\toprule

Model &  EM & B$_4$ & F$_1$  \\
\midrule
BART$_{\operatorname{base}}$ \citep{lewis-etal-2020-bart} &  70.1 & 83.9	& 69.5\\
QUEEN \citep{si-etal-2022-mining} & 71.6 & 86.3 & NA\\
SGT \citep{chen-2023-incomplete} & 71.1 & 86.7 & 85.0\\
MIUR \citep{li-etal-2023-well} & 70.9 & 86.0 & 72.3\\
Locate-Fill \citep{li-etal-2023-incomplete} & 75.0 & 87.3 & 84.2\\
XSS \citep{10384316} & 70.2 & 85.6 & 70.4\\
\midrule
TEO-Stage1 & 52.7 & 75.7 & 62.5\\
TEO-RFIS & 77.7 & 88.5 & 85.3\\
TEO-T5 & 76.4	& 87.6	& 84.5\\
TEO (Ours) & \textbf{78.1} & \textbf{88.7} & \textbf{85.8}\\
\midrule
TEO-Gold & 83.5 & 91.5 & 90.9\\
\bottomrule
\end{tabular}
\caption{Result comparison on English TASK.}
\label{tab:task}
\end{table}

\begin{table*}[t]
\centering
\small
\setlength{\tabcolsep}{3.6mm}
\begin{tabular}{lcccccccccccccc}
\toprule

Model &  F$_1$ & F$_2$ & F$_3$ & B$_1$ & B$_2$ & R$_1$ & R$_2$ & R$_L$ & EM  \\
\midrule

BART$_{\operatorname{base}}$ \citep{lewis-etal-2020-bart} & 81.2 & 76.0 & 79.7 & 93.9 & 90.8 & 95.2 & 91.8 & 92.4 & 70.5\\
RAU \citep{DBLP:conf/icassp/ZhangLWCX22} & NA & NA & NA & NA & 91.6 & NA & 90.6 & 93.9 & 68.4\\
QUEEN \citep{si-etal-2022-mining} & NA & NA & NA & NA & 92.1 & NA & 90.9 & 94.6 & 70.1\\
MIUR \citep{li-etal-2023-well} & NA & 82.2 & NA & NA & 91.2 & NA & 90.7 & 93.7 & 67.7\\
Locate-Fill \citep{li-etal-2023-incomplete} & 89.9 & 83.9 & 79.4 & 93.8 & 91.8 & 95.9 & 91.6 & 94.0 & 70.9\\
XSS \citep{10384316} & 89.8 & 82.0 & 76.1 & 92.4 & 91.0 & 95.8 & 90.7 & 93.7 & 66.7\\
\midrule
TEO-Stage1 & 85.6 & 63.0 & 28.3 & 90.9 & 82.6 & 94.0 & 75.9 & 75.9 & 34.0\\
TEO-RFIS & 90.0 & 83.9 & 80.9 & 93.3 & 91.2 & 95.9 & 91.6 & 94.2 & 72.1\\
TEO-T5 & \textbf{91.2}	& \textbf{85.9}	& 81.1 & 94.1	& 92.2	& 96.2	& 92.0	& 94.1	& 72.6\\
TEO (Ours) & 91.0 & 85.4 & \textbf{82.1} & \textbf{94.4} & \textbf{92.5} & \textbf{96.3} & \textbf{92.3} & \textbf{94.7} & \textbf{73.5}\\
\midrule
TEO-Gold & 98.1 & 94.3 & 91.2 & 98.4 & 97.3 & 99.1 & 96.7 & 97.5 & 86.3\\
\bottomrule
\end{tabular}
\caption{Result comparison on Chinese REWRITE.}
\label{tab:rewrite}
\vspace{-0.1cm}
\end{table*}

\begin{table*}[t]
\centering
\small
\setlength{\tabcolsep}{1.5mm}
\begin{tabular}{lcccccccccccccc}
\toprule

Model &  P$_1$ & R$_1$ & F$_1$ & P$_2$  & R$_2$ & F$_2$  & P$_3$  & R$_3$ & F$_3$ & B$_1$ & B$_2$ & R$_1$ & R$_2$ & EM \\
\midrule
BART$_{\operatorname{base}}$ \citep{lewis-etal-2020-bart} &  70.9 & 55.8	& 62.4	& 60.8	& 47.4	& 53.3	& 54.0	& 41.8	& 47.1	& 90.5	& 87.9	& 91.8	& 85.5 & 52.9\\
RAU \citep{DBLP:conf/icassp/ZhangLWCX22} & 75.0 & 65.5 & 69.9 & 61.2 & 54.3 & 57.5 & 52.5 & 47.0 & 49.6 & 92.4 & 89.6 & 92.8 & 86.0 & NA\\
QUEEN \citep{si-etal-2022-mining} & NA & NA & NA & NA & NA & NA & NA & NA & NA & 92.4 & 89.8 & 92.5 & 86.3 & 53.5\\
MGIIF \citep{du-etal-2023-multi} & NA & NA & 70.8 & NA & NA & 58.5 & NA & NA & 50.5 & \textbf{93.1} & \textbf{90.4} & 93.2 & 86.6 & NA\\
MIUR \citep{li-etal-2023-well} & 76.4 & 63.7 & 69.5 & 62.7 & 52.7 & 57.3 & 54.3 & 45.9 & 49.7 & 93.0 & 90.1 & 92.6 & 85.7 & 51.0\\
Locate-Fill \citep{li-etal-2023-incomplete} & 73.1 & 61.9 & 67.0 & 62.6 & 52.4 & 57.0 & 55.4 & 46.0 & 50.2 & 92.5 & 89.9 & 92.5 & 86.3 & 53.6\\
XSS \citep{10384316} & NA & NA & \textbf{70.9} & NA & NA & 57.0 & NA & NA & 47.9 & 92.5 & 89.7 & 92.7 & 85.9 & 50.1\\
\midrule
TEO-Stage1 & 73.8 & 65.0 & 69.1 & 42.7 & 39.5 & 41.0 & 24.6 & 23.0 & 23.7 & 92.9 & 86.8 & 92.6 & 78.6 & 42.5 \\
TEO-RFIS & 74.2 & 66.6 & 70.2 & 62.7 & 55.9 & 59.1 & 55.0 & 48.9 & 51.8 & 92.5 & 89.7 & 92.7 & 86.2 & 53.3\\
TEO-T5 & \textbf{76.7}	& 65.6	& 70.7	& 61.4	& 55.1	& 58.1	& 53.8	& 48.3	& 50.9	& 91.3	& 88.7	& \textbf{93.6}	& \textbf{87.9} & 53.6 \\
TEO (Ours) & 74.4 & \textbf{66.7} & 70.3 & \textbf{63.4} & \textbf{56.1} & \textbf{59.5} & \textbf{55.9} & \textbf{49.1} & \textbf{52.3} & 92.8 & 90.1 & 92.8 & 86.5 & \textbf{54.2}\\
\midrule
TEO-Gold & 93.5 & 94.3 & 93.9 & 83.9 & 84.1 & 84.0 & 76.8 & 76.9 & 76.9 & 97.6 & 95.6 & 98.0 & 93.6 & 80.3 \\
\bottomrule
\end{tabular}
\caption{Result comparison on Chinese RES200K.}
\label{tab:restoration}
\vspace{-0.1cm}
\end{table*}

\start{Datasets}
We conduct experiments on three popular IUR datasets: the Chinese open-domain dialogue datasets REWRITE \citep{su-etal-2019-improving} and RESTORATION-200K (abbreviated as RES200K) \citep{pan-etal-2019-improving}, and the English task-oriented dataset TASK \citep{quan-etal-2019-gecor}. Specific statistics for the datasets are provided in Appendix~\ref{appendix:dataset}.

\start{Metrics} 
In order to evaluate our method, we employ four different automatic evaluation metrics, namely EM, BLEU$_n$ (abbreviated as B$_n$) \citep{papineni-etal-2002-bleu}, ROUGE$_n$ (R$_n$) \citep{lin-2004-rouge}, and Restoration F-score$_n$ (F$_n$) \citep{pan-etal-2019-improving}. 
These metrics can effectively reflect the quality of rewriting. Following previous work \citep{liu-etal-2020-incomplete,si-etal-2022-mining, li-etal-2023-well}, we also report different metrics on different datasets.

\start{Baselines} We compare our TEO with the following strong baselines: 
BART$_{\operatorname{base}}$ \citep{lewis-etal-2020-bart}, 
QUEEN \citep{si-etal-2022-mining}, RAU \citep{DBLP:conf/icassp/ZhangLWCX22}, SGT \citep{chen-2023-incomplete}, MIUR \citep{li-etal-2023-well}, Locate-Fill \citep{li-etal-2023-incomplete}, MGIIF \citep{du-etal-2023-multi} and XSS \citep{10384316}.

The implementation and details of metrics and baselines are provided in Appendix~\ref{implementation},~\ref{d_mt} and~\ref{d_bs}.

\subsection{Experimental Results}
We evaluate our TEO and the baselines on the three datasets and the results are shown in Tables  \ref{tab:task}, \ref{tab:rewrite} and \ref{tab:restoration}, respectively, where “NA” refers to the metrics that the original paper did not report. The results on three datasets show that our TEO outperforms the baselines on almost all metrics. In terms of the most rigorous metric, EM, our TEO achieves an improvement of 3.1, 2.6 and 0.6 on the TASK, REWRITE and RES200K datasets, respectively, in comparison with the previous SOTA models. This result demonstrates that the robustness of our TEO allows it to avoid various minor boundary errors. Furthermore, the improvements observed on the English TASK and Chinese REWRITE and RES200K datasets demonstrate the applicability of our TEO to diverse languages and domains.

Furthermore, it is observed that on the three datasets, our TEO exhibits superior performance compared to the baselines in F$_n$, while it is almost on par with them in BLEU$_n$. This can be attributed to the fact that F$_n$ emphasizes the ability of the model to identify omitted or referential information in the incomplete utterance, whereas BLEU$_n$ considers all n-gram information present in the utterance including the span that is already present in the original incomplete utterance. The higher performance in F$_n$ indicates that our TEO can better capture contextual information of conversations and subsequently address  the issues of coreference and ellipsis based on the context. 
Moreover, we find that our TEO achieves the improvements of 0.8, 2.2 and 2.6 in F$_1$, F$_2$, and F$_3$ scores, respectively, in comparison with the previous SOTA model MIUR on the RES200K dataset. 
In essence, the F$_1$ score is indicative of the accuracy of token restoration, irrespective of the insertion position within the utterance. However, the positioning of newly added tokens will influence the values of  F$_2$ and F$_3$. The enhanced performance observed in  F$_2$ and F$_3$ also suggests that our TEO is capable of not only generating missing tokens but also inserting them at the appropriate locations.

To ensure a fair comparison, we maintained consistent experimental settings with previous related studies and utilized BART$_{\operatorname{base}}$ as the backbone. Additionally, we conducted experiments on three datasets using T5$_{\operatorname{base}}$ as the backbone and the results of these experiments are presented in Tables \ref{tab:task}, \ref{tab:rewrite} and \ref{tab:restoration}, as TEO-T5, respectively. It is worth noting that BART$_{\operatorname{base}}$ outperformed T5$_{\operatorname{base}}$ in most metrics, which may be attributed to the different pre-training objectives of the two models. During pre-training, BART introduces noise to the text and reconstructs the original text at the decoder. In contrast, T5 models various classification and generation tasks in a unified text-to-text format during pre-training. BART's pre-training objective is similar to our IUR task because coreference and ellipsis in IUR can be viewed as a type of noise that our task aims to recover.

\subsection{Analysis on Editing Operation Generation}
As shown in Table \ref{tab:stage1_em}, we calculate the EM metric of the editing operations generated in the first stage, i.e., editing operation generation. It can be observed that the EM metric  is also relatively high. For example, in the TASK and REWRITE datasets, the EM metrics of the first stage are 73.1 and 75.2, respectively, while the metrics for the second stage are 78.1 and 73.5, respectively. It can be observed that the EM metric is higher in the second stage of the TASK dataset in comparison with the first stage.  This further corroborates the hypothesis that even if erroneous editing operations are generated in the first stage, a portion of them can be rectified in the second stage. 

To further validate the effectiveness of the first stage, we use the correct editing operations during inference in the second stage, shown in Tables \ref{tab:task}, \ref{tab:rewrite} and \ref{tab:restoration}, as TEO-Gold, respectively.
The results show a significant improvement in all metrics for all three datasets after using the correct editing operations. This indicates that the performance of editing operation generation is positively correlated with that of the second stage. Simultaneously, the performance of directly using BART$_{\operatorname{base}}$, i.e., removing the first stage, is much lower than our TEO, which also proves the effectiveness of the first stage  and illustrates that the editing operations are the pivot for the IUR task. 

\subsection{Analysis on Editing-aware Rewritten Utterance Generation}

To verify the effectiveness of the second stage, i.e., editing-aware rewritten utterance generation, it is not necessary to proceed to the second stage after completing the first stage. Instead, the rewritten utterance can be generated directly according to the following rules: For those replacement operations, we directly parse the editing operations generated in the first stage. For insertion operations, since it is not possible to determine the exact positions at which the insertion should occur, a random selection of positions is made.  The results are shown in Tables \ref{tab:task}, \ref{tab:rewrite} and \ref{tab:restoration}, as TEO-Stage1 respectively. It can be observed that the outcomes of the first stage are considerably inferior to those of the two-stage method TEO. This evidence corroborates the efficacy of the second stage and the beneficial interaction between the first and second stages.

\begin{table}[t]
    \centering
    \small
\setlength\tabcolsep{13pt}
    \begin{tabular}{cccc}
    \toprule
    Dataset & EM & E2C(\%) & C2E(\%) \\
    \midrule
       TASK  & 73.1 & 9.71 & 20.56 \\
       REWRITE  & 75.2 & 12.53 & 3.41 \\
       RES200K & 59.8 & 4.82 & 29.66 \\
    \bottomrule
    \end{tabular}
    \caption{The EM, E2C and C2E of the first stage.}
    \label{tab:stage1_em}
    \vspace{-0.4cm}
\end{table}

In addition, we define two metrics, error to correct rate ($E2C$) and correct to error rate ($C2E$),
\begin{equation}
\label{equation:ecce}
    \small E2C = \frac{\#err\_cor}{\#er},
    \small C2E = \frac{\#cor\_err}{\#cor},
\end{equation}
where $\#err\_cor$ refers to the number of samples that were predicted incorrectly in the first stage but correctly in the second stage, $\#er$ refers to the number of samples that were predicted incorrectly in the first stage, $\#cor\_err$ refers to the number of samples that were predicted correctly in the first stage but incorrectly in the second stage, and $\#cor$ refers to the number of samples that were predicted correctly in the first stage. The first metric measures the proportion of samples that were incorrectly predicted in the first stage but correctly predicted in the second stage. The second metric measures the proportion of samples that were correctly predicted in the first stage but incorrectly predicted in the second stage. With these two metrics, we can quantitatively analyze the correlation between the two stages.

As shown in Table \ref{tab:stage1_em}, we find that $E2C$ is relatively higher than $C2E$ on the REWRITE dataset. This indicates that a higher proportion of incorrect editing operations in the first stage are corrected in the second stage, thereby proving that the combination of global dialogue context information with local editing operation information is effective. Through our observations, we find that there are far more cases of ellipsis than the cases of coreference  in RES200K and TASK. This implies that there are more instances of insertion. In the first stage, we only obtain the inserted tokens but do not know its positions. Therefore, even if we correctly predict the inserted tokens in the first stage, there are still many cases of incorrect insertion positions in the second stage. Consequently, the value of $C2E$ is higher in RES200K  and TASK.

Since the first stage  can solve the issue of coreference, the replacement operations can be used to rewrite utterances. Instead of that, we feed  both the replacement and insertion operations to the second stage. To conduct insightful analysis, we perform predictive replacement operations after the first stage and only handle insertion operations in the second stage. The results are shown in Tables \ref{tab:task}, \ref{tab:rewrite}, \ref{tab:restoration} as TEO-RFIS, respectively.  This approach yields inferior results across all metrics compared to simultaneously handling replacement and insertion operations in the second stage. Taking table \ref{case_stage} as example, the first stage generates an incorrect replacement operation, predicting the replacement of ``this musical instrument'' with ``this''. If the replacement is executed directly, an incorrect output is produced. However, the second stage is able to correct this error and produce the correct utterance.

\begin{table}[t]
\centering
\resizebox{\linewidth}{!}{
\begin{tabular}{p{1.21\linewidth}}
\toprule
\textbf{Context}: \\
$\mathbf{Speaker_1}$: Don't you have any musical instruments that you want to learn? I think the piano and guitar sound great. \\
$\mathbf{Speaker_2}$: Piano. \\
$\mathbf{Speaker_1}$: The sound of this musical instrument sounds very pleasant, wow. (\textbf{Incomplete utterance})  \\
\midrule
\textbf{Reference}: The sound of the piano sounds very pleasant, wow. \\
\midrule
\textbf{Correct editing operation}: \\\
 [D] this musical instrument [R] the piano \\
\midrule
\textbf{Predicted editing operation in the first stage}: \\\
 [D] this [R] the piano \xmark \\
\midrule
\textbf{Predicted rewritten utterance in the second stage}: \\
The sound of the piano sounds very pleasant, wow. \cmark \\
\bottomrule
\end{tabular}
}
\caption{
A boundary error in the first stage is corrected by the second stage.}
\label{case_stage}
\vspace{-0.3cm}
\end{table}

\begin{figure}[t]
\begin{center}
 \includegraphics[width=1\linewidth]{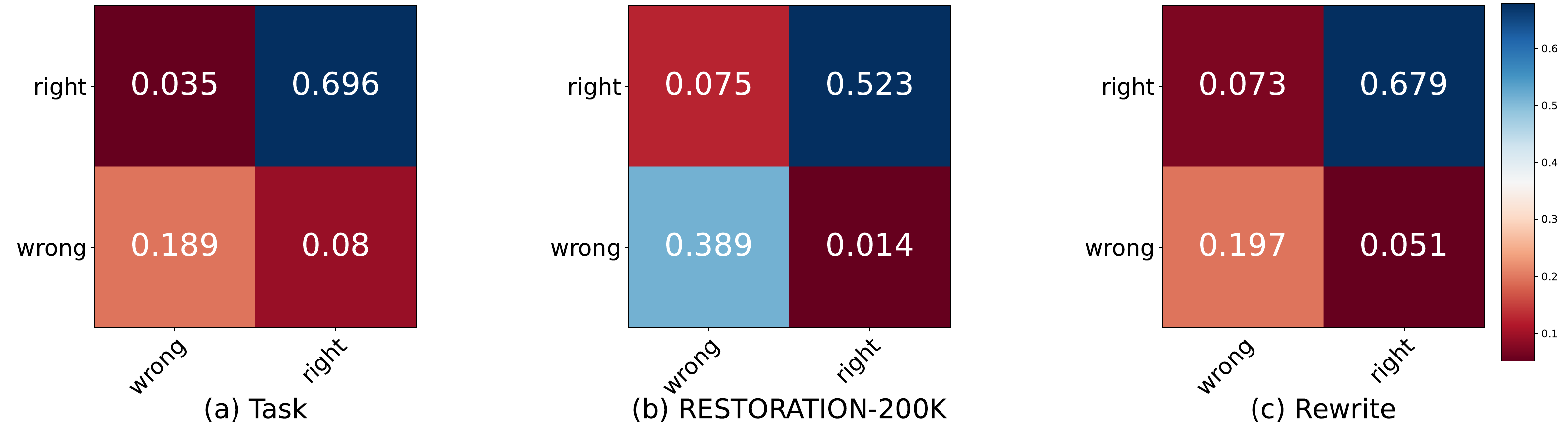}
 \caption{The correspondence between the positive and negative examples in the first stage and the second stage, where 
 ``wrong'' and ``right'' on the vertical axis respectively represent incorrect and correct editing operations in the first stage, while ``wrong'' and ``right'' on the horizontal axis respectively represent incorrect and correct rewritten utterances in the second stage.}
 \label{fig:comp}
\end{center}
 \vspace{-0.6cm}
\end{figure}

Furthermore, we also investigated the results predicted in the second stage for the samples corresponding to correct or incorrect editing operations predicted in the first stage. The results are shown in Figure \ref{fig:comp}.  The majority of samples with accurate editing operation predictions in the first stage were correctly identified in the second stage, while some of the incorrectly predicted samples were mitigated in the second stage. We find that 
8.0\%, 5.1\% and 1.4\% of samples are corrected by the second stage on TASK, REWRITE, and RES200K respectively. However, these samples did not have correct editing operations in the first stage. This suggests that our second stage editing-aware rewritten utterance generation is capable of correcting the errors produced in the first stage.

It is worth noting that the edit operations predicted in the first stage can be used in the training of the second stage and our experimental results show the values of F$_1$ and EM were 89.7 and 72.3, respectively, which are inferior to our TEO (91.2 and 73.5). This is because it would result in a fixed distribution of erroneous editing operations in the training data of the second stage. Nevertheless, the utilization of our proposed adversarial perturbation strategy enables the dynamic adjustment of noise within training samples, thereby enhancing the model's robustness. 

\begin{figure}[t]
\begin{center}
 \includegraphics[width=1\linewidth]{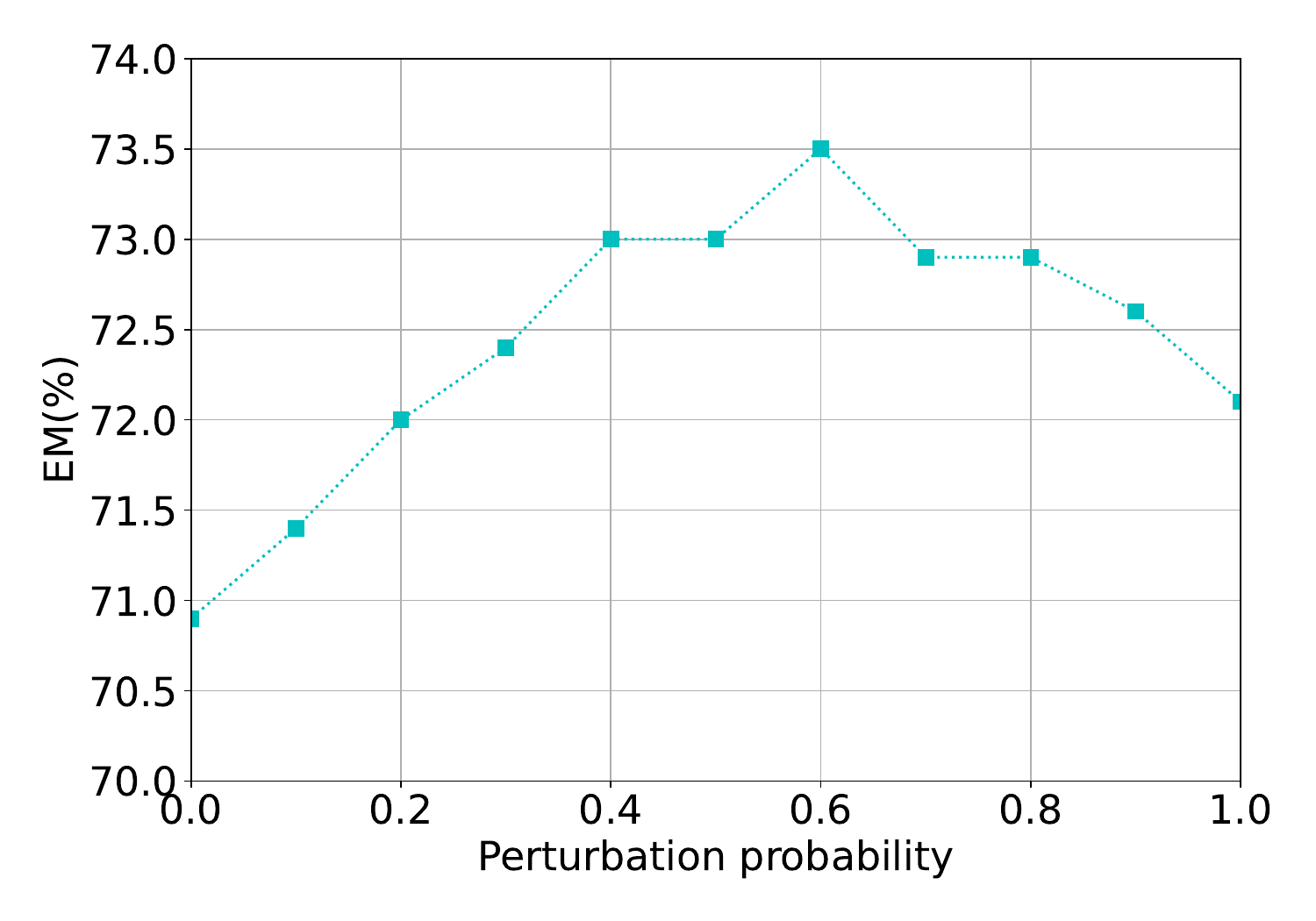}
 \caption{The trend chart of EM metric changing with the variation of adversarial perturbation probability on REWRITE.
 }
 \label{fig:perturb}
\end{center}
\vspace{-0.7cm}
\end{figure}

\subsection{Impact of Adversarial Perturbation}
\label{appendix:adv}
As mentioned in Section \ref{sec:editgen},  the introduction of adversarial perturbations is employed to mitigate the impact of exposure bias, which arises due to inconsistency between the training and prediction stages. To assess the efficacy of adversarial perturbations, experiments were conducted with varying perturbation probabilities, and the experimental results are presented in Figure \ref{fig:perturb}. As the probability of perturbation increases, the EM value initially increases and then decreases, reaching its peak when the probability is 0.6. This may be due to the inconsistency between the training and inference data distributions when the perturbation probability is low. When the perturbation probability is too high, TEO fails to capture knowledge in editing operations. However, we observe that even when the perturbation probability is 1, TEO can still achieve good results. This also indicates that in situations that are close to zero-shot, our TEO can still perform well.

\begin{figure}[t]
\begin{center}
 \includegraphics[width=1\linewidth]{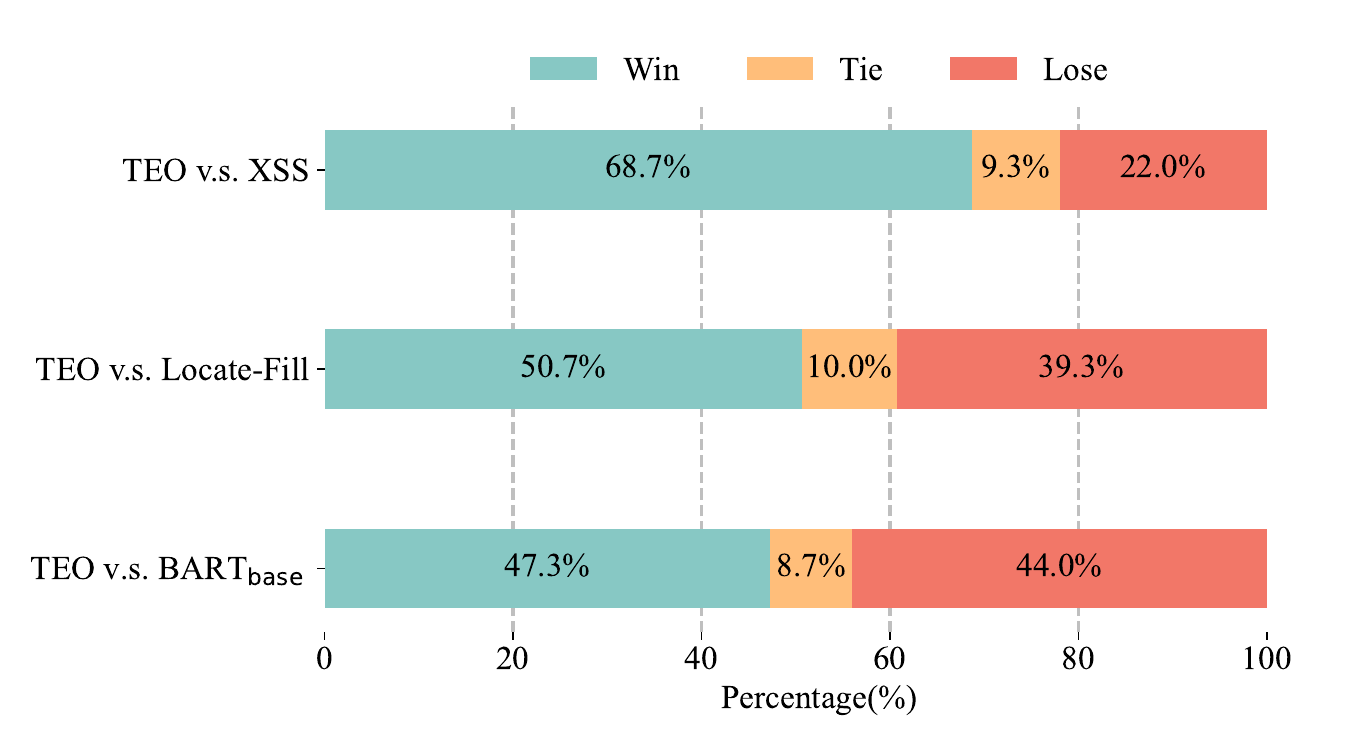}
 \caption{Performance comparison of our method with BART$_{\operatorname{base}}$, Locate-Fill and XSS on human evaluation.}
 \label{fig:h_eval}
\end{center}
 \vspace{-0.6cm}
\end{figure}

\subsection{Human Evaluation}
Incomplete utterance rewriting often results in non-unique rewriting outcomes, making it challenging to fully assess the performance of the model based on automatic evaluation metrics alone. To address this, we conducted a human evaluation to compare our proposed method with BART$_{\operatorname{base}}$, Locate-Fill and XSS.  A total of fifty data points were randomly sampled from the test set and distributed to three raters. Each rater selected a better result from the outputs generated by two methods. 

The results are presented in Figure~\ref{fig:h_eval}, it can be seen that our method outperforms BART$_{\operatorname{base}}$, Locate-Fill and XSS, indicating consistency with the results observed in our automatic evaluation. However, it was observed that our method has a smaller advantage when evaluated manually against BART$_{\operatorname{base}}$, compared to Locate-Fill and XSS. After analyzing the rewritten utterances produced by these four methods, it was discovered that while BART$_{\operatorname{base}}$ demonstrated an average ability to rewrite, it was capable of generating more fluent utterances, which ultimately led to superior human evaluation results.

\begin{table}[t]
\centering
\resizebox{\linewidth}{!}{
\small
\begin{tabular}{lccccccccccccc}
\toprule

Model & F$_1$ & F$_2$  & B$_1$ & B$_2$ & R$_1$ & R$_2$ & R$_L$ & EM   \\
\midrule

GPT-4 & 79.1	& 66.7	& 85.3	& 80.4	& 87.7	& 77.6	& 82.1	& 35.7\\
\midrule
Ours & \textbf{91.0} &	\textbf{85.4} &	\textbf{94.4} &	\textbf{92.5} &	\textbf{96.3} &	\textbf{92.3} &	\textbf{94.7} &	\textbf{73.5}\\
\bottomrule
\end{tabular}
}
\caption{Performance comparison between TEO and GPT-4 on REWRITE.}
\label{tab:gpt-comp}
\vspace{-0.4cm}
\end{table}

\subsection{Comparison with ChatGPT}
At present, the majority of LLMs lack the capacity to process coreference and ellipsis resolution, which are integral to IUR. We conducted preliminary experiments utilising the in-context learning method with GPT-4 (the version used here is \texttt{gpt-4-1106-preview}) as a case study and the prompt used in GPT-4 is provided in Appendix~\ref{prompt}. 

We provided five samples as demonstrations to GPT-4 and the experimental results on REWRITE are shown in Table~\ref{tab:gpt-comp}. Even though the number of parameters in GPT-4 is much higher than our model, our model still achieves better performance than it. In addition, we observed that some utterances generated by GPT-4 do not match incomplete utterances in terms of semantics, which also indicates that the understanding of global conversational semantics by LLMs needs to be improved.

\subsection{Case Study}
\label{appendix:case}
Our TEO is capable of accommodating both instances where the rewritten utterance contains tokens that are not present in the dialogue history, as well as instances where multiple discontinuous spans must be inserted at the same position in the current utterance. We compare our TEO with three baselines, i.e., BART$_{\operatorname{base}}$, RUN, and HCT, and the results on the Chinese REWRITE dataset are shown in Table \ref{case_comp}. We need to insert four spans ``\begin{CJK*}{UTF8}{gbsn}天龙八部里\end{CJK*}'' (in Demi-Gods and Semi-Devils), ``\begin{CJK*}{UTF8}{gbsn}段誉\end{CJK*}'' (Duan Yu), ``\begin{CJK*}{UTF8}{gbsn}的\end{CJK*}'' (of), and ``\begin{CJK*}{UTF8}{gbsn}武功最高\end{CJK*}'' (the highest martial arts skill) at the end of the current utterance. Although BART$_{\operatorname{base}}$ generates a more fluent rewritten utterance, it does not fit into the context of the conversation. RUN selects all correct spans but fails to insert them in the correct order within the utterance (Although the English translation is coherent, there is an word order error in Chinese sentences). HCT has boundary errors for some spans; for example, instead of inserting ``\begin{CJK*}{UTF8}{gbsn}段誉\end{CJK*}'' (Duan Yu), it inserts ``\begin{CJK*}{UTF8}{gbsn}是段誉\end{CJK*}'' (is Duan Yu). Additionally, there is also an issue with span insertion order. Only our TEO outputs correct and complete utterance, which benefits from the complementary effects of the two-stage mechanism.

\begin{table}[t]
\centering
\resizebox{\linewidth}{!}{
\begin{tabular}{p{1.21\linewidth}}
\toprule
\textbf{Context}: \\
$A$: \begin{CJK*}{UTF8}{gbsn}\colorbox{gray!50}{天龙八部里}谁\colorbox{purple!50}{的}\colorbox{orange!50}{武功最高}\end{CJK*}(Who has the highest martial arts skill in ``Demi-Gods and Semi-Devils''?) \\
$B$: \begin{CJK*}{UTF8}{gbsn}是\colorbox{red!50}{段誉}\end{CJK*}(It's Duan Yu.) \\
$A$: \begin{CJK*}{UTF8}{gbsn}为什么\end{CJK*}(Why?) (\textbf{Incomplete utterance}) \\
\midrule
\textbf{Reference}: \begin{CJK*}{UTF8}{gbsn}为什么\colorbox{gray!50}{天龙八部里}\colorbox{red!50}{段誉}\colorbox{purple!50}{的}\colorbox{orange!50}{武功最高}\end{CJK*}(Why is Duan Yu's martial arts the highest in ``Demi-Gods and Semi-Devils''?) \\
\midrule
\textbf{BART$_{\operatorname{base}}$}: 
 \begin{CJK*}{UTF8}{gbsn}为什么天龙八部里谁的武功最高(Why is whose  martial arts the highest in ``Demi-Gods and Semi-Devils''?) \end{CJK*} \xmark \\
\midrule
\textbf{RUN}: 
\begin{CJK*}{UTF8}{gbsn}为什么天龙八部里的武功最高段誉(Why is the martial arts in ``Demi-Gods and Semi-Devils'' the highest, Duan Yu?)\end{CJK*} \xmark\\
\midrule
\textbf{HCT}: 
 \begin{CJK*}{UTF8}{gbsn}为什么天龙八部里谁武功最高是段誉(Why is Duan Yu whose martial arts is the highest in ``Demi-Gods and Semi-Devils''?)\end{CJK*} \xmark \\
\midrule
\textbf{TEO (Ours)}: 
 \begin{CJK*}{UTF8}{gbsn}为什么天龙八部里段誉的武功最高(Why is Duan Yu's martial arts the highest in ``Demi-Gods and Semi-Devils''?)\end{CJK*} \cmark \\
\bottomrule
\end{tabular}
}
\caption{
A case of our TEO and three baselines BART$_{\operatorname{base}}$, RUN and HCT on REWRITE.
}
\vspace{-0.5cm}
\label{case_comp}
\end{table}

\subsection{Error Analysis}

To analyse the errors in our TEO model, we compiled the experimental results and found that the majority of errors arise from the insertion operation rather than the replacement operation. For instance, in REWRITE, insertion errors account for 79.3\%, while replacement errors account for only 20.7\%. This outcome is mainly due to the uncertainty of the insertion position. Secondly, in contrast to REWRITE, RES200K and TASK contain utterances that do not require rewriting. This can lead to errors when editing these utterances. For example, in RES200K, 38.7\% of utterances do not require rewriting and have an
EM score of 77.5, with almost all errors resulting from incorrect replacement operations. Finally, the EM scores for the replacement operation are considerably lower for RES200K (0) and TASK (28.6) than for REWRITE (83.0). This is primarily due to the limited number of replacement operations in these two datasets (RES200K: 0.15\%; TASK: 12.42\%).

\section{Conclusion}
\label{sec:bibtex}
We propose a two-stage IUR framework by taking the editing operations as the pivot, in which the first stage generates editing operations for IUR and the second stage rewrites incomplete utterances utilizing the generated editing operations and the dialogue context. Moreover, an adversarial perturbation strategy is proposed to enhance model robustness. The experimental results on three IUR datasets show that our TEO outperforms the SOTA models  significantly. Our future  work will focus on how to introduce LLMs to assist IUR.

\section*{Limitations}
Although this paper may contribute to incomplete utterance rewriting and some downstream dialogue tasks, it still suffers from two shortcomings, which are our future work. First, we only use one representation of the editing operation in our model. We believe that better templates can help the model better understand and improve the effectiveness of dialogue rewriting. Second, we only used the editing operations generated in the first stage to assist in rewriting the dialogue utterances in the second stage, but did not attempt to use the dialogue rewriting of the second stage to facilitate the first stage. Therefore, how to promote the complementary interaction between the two stages is our future research.

\section*{Acknowledgements}
The authors would like to thank the three anonymous reviewers for their comments on this paper. This research was supported by the National Natural Science Foundation of China (Nos. 62376181 and 62276177), and Project Funded by the Priority Academic Program Development of Jiangsu Higher Education Institutions.

\bibliography{custom}

\appendix

\section{Dataset Statistics}
\label{appendix:dataset}
The specific information of the three datasets is shown in Table ~\ref{tb:data_stat}.

\begin{table}[h]
\centering
\setlength\tabcolsep{8pt}
\small
\begin{tabular}{cccc}
\toprule
Category & REWRITE & RES200K & TASK \\
\midrule
Language & Chinese & Chinese & English \\
Train & 18K & 194K & 2.2K \\
Dev & 2K & 5K & 0.5K \\
Test & 2K & 5K & 0.5K \\
\#Avg. Cont & 17.7 & 25.5 & 52.6 \\
\#Avg. Curr & 6.5 & 8.6 & 9.4 \\
\#Avg. Rewr & 10.5 & 12.4 & 11.3 \\
\#Insertion & 14070 & 136339 & 1572 \\
\#Replacement & 7853 & 203 & 223 \\
\bottomrule
\end{tabular}
\caption{Statistics of different datasets, where  ``Cont'', ``Curr'' and ``Rewr'' are the abbreviations for the context, current, and rewritten utterance, respectively.}
\label{tb:data_stat}
\end{table}

\section{Implementation}
\label{implementation}
The pre-trained BART$_{\operatorname{base}}$ is employed as the backbone model, with all experiments conducted using the open-source library PyTorch. To reduce the latency during inference, a greedy decoding strategy is employed. Both the first stage and the second stage are fine-tuned for 30 epochs. The AdamW optimiser \citep{DBLP:conf/iclr/LoshchilovH19}  is used, with a learning rate of 5e-5. In the second stage, the probabilities of random deletion and random replacement are both set to 0.5. Given that the length of the edit operation is considerably shorter than that of the rewritten utterance, the decoding time of the first stage is negligible in comparison to that of the second stage.

\section{Details of Metrics}
\label{d_mt}

EM refers to exact matching accuracy, which is a strict metric, representing the ratio of correctly predicted samples to the total number of samples. The BLEU metric evaluates accuracy by calculating the matching degree of n-grams. BLEU$_1$  is a metric that measures the accuracy at the word level, while higher-order BLEU can be used to assess the fluency of utterances. Additionally, we employ ROUGE$_n$ to measure recall in IUR. ROUGE evaluates recall by counting n-gram co-occurrences. F$_n$ \citep{pan-etal-2019-improving} is utilized to identify the words that have been added to the utterance for rewriting. The n-gram restoration precision, recall and F-score are calculated as follows, 
\begin{equation}
    \begin{aligned}
    p_n & =\frac{\mid\{\text { restored n-grams }\} \cap\{\text { n-grams in ref }\} \mid}{\mid\{\text { restored n-grams }\} \mid} \\
    r_n & =\frac{\mid\{\text { restored n-grams }\} \cap\{\text { n-grams in ref }\} \mid}{\mid\{\text { n-grams in ref }\} \mid} \\
    f_n & =2 \cdot \frac{p_n \cdot r_n}{p_n+r_n}
    \end{aligned} \nonumber
\end{equation}
\noindent where ``restored n-grams'' denotes n-grams in model-generated utterances that contain at least one restored word, and ``n-grams in ref'' denotes n-grams in reference utterances that contain at least one restored word.

\section{Details of Baselines}
\label{d_bs}
We introduce eight strong baselines to verify the effectiveness of our proposed model TEO as follows.

(1) BART$_{\operatorname{base}}$: it generated rewritten utterance using dialog history and incomplete utterance as input; 

(2) QUEEN \citep{si-etal-2022-mining}: it proposed a query template that was concatenated with utterance as input; 

(3) RAU \citep{DBLP:conf/icassp/ZhangLWCX22}: it extracted relations between tokens from a self-attention matrix; 

(4) SGT \citep{chen-2023-incomplete}: it first identified fragments and their relative order, and then generated the target utterance; 

(5) MIUR \citep{li-etal-2023-well}: it mined latent semantic information through a layer of MLP and predicted token types through a joint feature matrix; 

(6) Locate-Fill \citep{li-etal-2023-incomplete}: it proposed a two-phase incomplete utterance rewriting method that first predicted empty slots and then filled them; 

(7) MGIIF \citep{li-etal-2023-incomplete}: it proposed a multi-task information interaction framework for incomplete utterance rewriting; 

(8) XSS \citep{10384316}: it is an incomplete utterance rewriting model based on span pairing.

\section{Prompts Used in GPT-4 Evaluation}
\label{prompt}
The prompt used in the GPT-4 evaluation is as follows.
\begin{tcolorbox}[colback=Emerald!9,colframe=cyan!40!black,title=Prompt used in GPT-4 assessment]
  The goal of dialogue rewriting is to resolve coreference and ellipsis, that is, to complete the coreferential and omitted information in the dialogue without changing its original semantics. Please rewrite the final utterance in the following dialogue. \\
  Examples: \{\texttt{Examples}\} \\
  Input: \{\texttt{Input}\}
\end{tcolorbox}

\end{document}